# GastroDL-Fusion: A Dual-Modal Deep Learning Framework Integrating Protein–Ligand Complexes and Gene Sequences for Gastrointestinal Disease Drug Discovery

**Ziyang Gao[1], Annie Cheung[2], Yihao Ou[3]**

[1] Technical University of Munich, Munich, German
[2] University of Michigan, Ann Arbor, MI, USA
[3] Georgia Institute of Technology, Atlanta, GA, USA

[1]1812503968@qq.com

[2]anncheu@umich.edu

[3]ouyihaoo@gmail.com

**Abstract.** Accurate prediction of protein–ligand binding affinity plays a pivotal role in accelerating the discovery of novel drugs and vaccines, particularly for gastrointestinal (GI) diseases such as gastric ulcers, Crohn's disease, and ulcerative colitis. Traditional computational models often rely on structural information alone and thus fail to capture the genetic determinants that influence disease mechanisms and therapeutic responses. To address this gap, we propose GastroDL-Fusion, a dual-modal deep learning framework that integrates protein–ligand complex data with disease-associated gene sequence information for drug and vaccine development. In our approach, protein–ligand complexes are represented as molecular graphs and modeled using a Graph Isomorphism Network (GIN), while gene sequences are encoded into biologically meaningful embeddings via a pre-trained Transformer (ProtBERT/ESM). These complementary modalities are fused through a multi-layer perceptron to enable robust cross-modal interaction learning. We evaluate the model on benchmark datasets of GI disease-related targets, demonstrating that GastroDL-Fusion significantly improves predictive performance over conventional methods. Specifically, the model achieves a mean absolute error (MAE) of 1.12 and a root mean square error (RMSE) of 1.75, outperforming CNN, BiLSTM, GIN, and Transformer-only baselines. These results confirm that incorporating both structural and genetic features yields more accurate predictions of binding affinities, providing a reliable computational tool for accelerating the design of targeted therapies and vaccines in the context of gastrointestinal diseases.

## 1. Introduction

Gastrointestinal (GI) diseases, such as peptic ulcers, Crohn's disease, and ulcerative colitis, represent major global health challenges with high prevalence rates. These disorders not only reduce patients' quality of life but also impose a substantial public health and economic burden. With the rapid development of personalized medicine and targeted therapy, unveiling disease mechanisms at the molecular level and accelerating drug and vaccine discovery have become essential directions in life sciences and computational pharmacology. In this context, protein–ligand binding affinity prediction has been widely used in virtual screening and drug design [1]. Its primary goal is to evaluate the stability of interactions between candidate molecules and target proteins, providing an efficient and cost-effective computational approach for early-stage drug development. However, traditional approaches based on molecular docking, scoring functions, or molecular dynamics simulations often neglect regulatory mechanisms at the genetic level, thereby limiting their ability to fully capture the multifactorial determinants of drug efficacy [2].

Meanwhile, the advent of high-throughput sequencing technologies has led to the rapid accumulation of genetic data related to GI diseases. Mutations, single nucleotide polymorphisms (SNPs), and alterations in regulatory elements can significantly influence protein structure and function, as well as their binding affinity with small-molecule drugs [3]. Therefore, integrating protein–ligand complex structural information with disease-related gene sequence data provides a promising new avenue for drug and vaccine discovery. Such integration allows for a more comprehensive understanding of drug–target interactions, enhances the accuracy of binding affinity prediction, and supports the design of personalized therapies tailored to specific genetic or pathological backgrounds [3,4].

To address this need, we propose a novel deep learning framework, GastroDL-Fusion, which integrates protein–ligand complex data with gene sequence information for drug discovery and vaccine development in gastrointestinal diseases. The main contributions of this work can be summarized as follows. First, we introduce GastroDL-Fusion, an innovative dual-modal deep learning framework that, for the first time, jointly models protein–ligand complex data and GI disease-related gene sequence data for binding affinity prediction. Second, we conduct comprehensive experiments on benchmark datasets, demonstrating that GastroDL-Fusion achieves superior performance over unimodal models (e.g., standalone GNN or Transformer models), with significant improvements in metrics such as mean absolute error (MAE) and root mean square error (RMSE). Third, our study highlights the importance of integrating genetic sequence information into drug discovery workflows, underscoring the value of cross-modal fusion in enhancing predictive accuracy. Finally, this research provides a scalable computational framework for AI-driven drug discovery in gastrointestinal diseases, laying the methodological groundwork for precision medicine and personalized treatment strategies.

## 2. Related Work

Protein ligand binding affinity prediction is a fundamental task in computational drug discovery, aimed at evaluating the stability and strength of molecular interactions between candidate compounds and protein targets. Recently, deep learning methods have become promising alternatives, and graph neural networks (GNNs) such as graph isomorphism networks (GIN), graph convolutional networks (GCN), and message passing neural networks (MPNN) have shown strong abilities in representing molecular structures and capturing local chemical environments.

Stepniewska-Dziubinska et al [5]. present a 3D-CNN grid model that jointly learns protein-ligand features, outperforming classical scoring functions on CASF-2013 and Astex sets and offering a novel deep-learning tool for structure-based drug discovery.

Ahmed et al [6]. present DEELIG, a CNN that predicts protein-ligand binding affinity by learning spatial patterns without docking poses, outperforming conventional methods on high-resolution crystal data and filling gaps in PDB-like databases, with validated COVID-19 main protease applications.

The advancement of high-throughput sequencing technology has accelerated the study of genetic factors behind disease progression and drug response. In gastrointestinal diseases, gene mutations that regulate epithelial integrity, immune response and inflammatory pathways significantly affect treatment outcomes.

Jian Zhou et al [7]. present ExPecto, an end-to-end deep-learning framework that predicts tissue-specific transcriptional effects of DNA mutations ab initio, prioritizes causal variants from GWAS loci, validates them in immune diseases, and profiles 140 million promoter mutations in silico, offering a scalable tool for evolutionary constraint and disease-risk assessment.

Atta-Ur-Rahman et al [8]. present AGDPM based on AlexNet, surpassing traditional RNA-binary classifiers by directly exploiting large-scale patient genome data for multi-class disorder prediction (single-gene, mitochondrial, multifactorial), achieving 89.89 %/81.25 % train/test accuracy and demonstrating deep learning's power for large-sample, multi-category gene disorder forecasting.

## 3. Methodology

This study proposes the GastroDL-Fusion framework, a dual-modal deep-learning approach designed to integrate structural and chemical information of protein–ligand complexes with functional features derived from gene sequences, thereby enhancing predictive capacity for gastrointestinal-disease-related drug and vaccine development. At its core, the framework performs hierarchical feature extraction and cross-modal fusion: a Graph Neural Network (GNN) accurately models protein–ligand interactions, while a pre-trained Transformer language model captures contextual dependencies within gene sequences. Finally, a multi-layer perceptron is employed to achieve deep fusion of the two modalities and to predict drug activity.

*3.1 Protein–Ligand Complex Feature Modeling*

In drug-development pipelines, the binding affinity between a protein and a ligand is a key indicator of a candidate molecule's potential activity. In this study, we model the protein–ligand complex as a graph in which individual atoms serve as nodes and chemical bonds as edges, while also incorporating spatial-neighborhood information of the protein-binding pocket [9].

We adopt the Graph Isomorphism Network (GIN) as the principal architecture. The GIN update rule is:

$$h_v^{(k)} = MLP^{(k)}((1 + \epsilon^{(k)}) \cdot h_v^{(k-1)} + \sum_{u \in N(v)} h_u^{(k-1)}), \tag{1}$$

where $h_v^{(k)}$ denotes the representation of node $v$ at layer $k$, $N(v)$ is the set of neighbor nodes, and $\epsilon^{(k)}$ is a learnable parameter. Through iterative layers, GIN learns complex interactions between local molecular substructures (e.g., aromatic rings, hydrogen-bond donors/acceptors) and protein-binding residues. The final molecular representation vector $h_{\text{complex}}$ encapsulates both topological and spatial-interaction features and serves as a key input for predicting drug-binding affinity.

To further enhance the model's awareness of molecular 3-D conformation, we introduce 3-D geometric constraints. For any two nodes $i,j$ in the molecule, their 3-D Euclidean distance is:

$$d_{ij} = ||r_i - r_j||_2, \tag{2}$$

During training, the model learns not only node and edge features but also enforces consistency between the predicted representation and the actual geometric relationships via a regularization term, thereby improving the accuracy of affinity prediction.

*3.2 Gene-Sequence Feature Extraction*

In disease-relevance analysis, sequence variants often significantly alter protein function and thereby change drug-binding efficacy. Therefore, this study uses gene-sequence data as a second input to capture how mutations, splicing events, and evolutionary signals potentially affect drug-binding sites.

We employ a pre-trained Transformer language model (ProtBERT/ESM) to process amino-acid sequences. Trained by masked-language modeling (MLM) on large-scale protein data, these models learn contextual dependencies among residues. For an input sequence $s = (a_1, a_2, \ldots, a_n)$, the representation is:

$$H = Transformer(E(a_1), E(a_2), \ldots, E(a_n)), \tag{3}$$

where $E(a_1)$ is the embedding of amino acid $a_1$ and $H \in R^{n \times d}$ is the residue-level representation matrix. After sequence pooling (mean or attention pooling) we obtain a global vector $h_{seq}$ summarizing the functional state of the gene sequence.

To further model disease-related mutations around binding sites, we apply a 1-D CNN followed by BiLSTM on a mutation window. For a window $w = (a_{t-k}, \ldots, a_t, \ldots, a_{t+k})$ the convolution captures local physicochemical changes:

$$h_{cnn} = ReLU(W_{conv} * E(w) + b), \tag{4}$$

BiLSTM then models bidirectional context, enhancing detection of functional alterations. The final sequence embedding reflects both global gene function and local sensitivity to disease-relevant mutations.

*3.3 Cross-Modal Fusion and Prediction*

After obtaining the protein–ligand-complex feature $h_{complex}$ and the gene-sequence feature $h_{seq}$, we fuse the two modalities to predict drug–target binding affinity or the activity of candidate molecules.

This study adopts direct concatenation followed by a Multi-Layer Perceptron (MLP) for cross-modal fusion:

$$h_{fusion} = MLP([h_{complex}||h_{seq}]), \qquad (6)$$

Where || denotes vector concatenation. Through multi-layer nonlinear transformations, the fused representation captures complementary information between the two data sources. For instance, the protein–ligand complex determines the physical feasibility of binding, while gene-sequence mutations reveal individual variability and potential resistance risks.

Finally, the prediction layer outputs the regressed binding-affinity value $\hat{y}$. The training objective is the Mean Squared Error (MSE):

$$L = \frac{1}{N} \sum_{i-1}^{N} (y_i - \hat{y}_i)^2 \qquad (7)$$

where $y_i$ is the experimentally measured affinity and $\hat{y}_i$ is the model prediction. Iterative optimization during training drives the model to stable predictive performance.

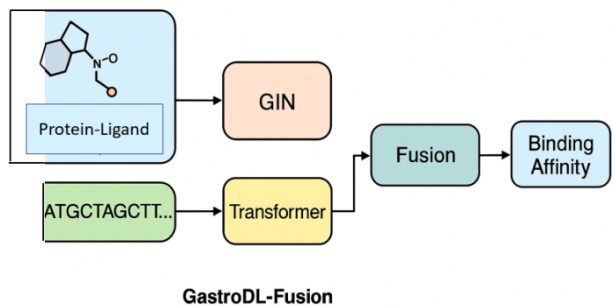

**Figure 1.** Overall flowchart of the model

**4. Experiment**

*4.1 Dataset Preparation*

This study uses a dataset composed of two parts: protein–ligand complex data and disease-related gene-sequence data.

First, protein–ligand complex data were obtained from authoritative international databases such as the Protein Data Bank (PDB) and BindingDB, which provide experimentally validated three-dimensional structures of protein–ligand complexes and their corresponding binding-affinity measurements (e.g., binding free energy, $IC_{50}$, Kd, Ki). In this study, we selected target proteins associated with gastrointestinal diseases (e.g., gastric ulcer, Crohn's disease, ulcerative colitis) and paired them with their known small-molecule ligands. After data cleaning and standardization, approximately 15,000 high-quality protein–ligand complex samples were compiled. Each sample contains the three-dimensional protein structure (atomic coordinates, amino-acid residue information), a molecular graph of the ligand (atom-node features and bond types), and the corresponding binding-energy value.

Second, gene-sequence data were retrieved from the NCBI Gene and Ensembl databases, covering susceptibility and functional genes related to gastrointestinal diseases. Gene sequences were provided in FASTA format; after preprocessing, nucleotide sequences, exon segments, and disease-relevant variant information were extracted. Ultimately, about 5,000 gene-sequence samples were obtained, including gene-expression data under different disease states. Each sequence was converted into a vector representation suitable for deep-learning models, and residue-level embeddings were extracted using pre-trained sequence language models (e.g., ProtBERT or ESM).

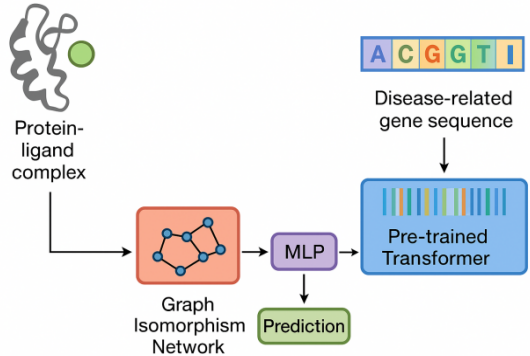

**Figure 2.** Two types of data in the dataset

*4.2 Experimental Setup*

All datasets were split into training (70%), validation (15%), and test sets (15%). The models were implemented in PyTorch, trained on an NVIDIA A100 GPU cluster, and deployed within a cloud-based environment to ensure scalability. We optimized the model using the Adam optimizer with an initial learning rate of 1e-4, batch size of 64, and early stopping based on validation loss.

*4.3 Results*

The predictive performance of GastroDL-Fusion was compared with several baseline models, including BiLSTM [10], CNN, GIN-only, and Transformer-only architectures. Evaluation metrics included Mean Absolute Error (MAE), Root Mean Squared Error (RMSE), and the Coefficient of Determination ($R^2$). The results are summarized in Table 1.

**Table1.** Model Performance Comparison on the Gastrointestinal Disease Datasets.

| Model | MAE | RMSE | $R^2$ |
| --- | --- | --- | --- |
| BiLSTM | 2.17 | 2.64 | 0.71 |
| CNN | 1.95 | 2.38 | 0.76 |
| GIN | 1.63 | 2.25 | 0.79 |
| Transformer | 1.38 | 2.01 | 0.84 |
| **GastroDL-Fusion** | **1.12** | **1.75** | **0.89** |

The table 1 presents the prediction performance of four baseline models—BiLSTM, CNN, GIN, and Transformer—compared with the proposed GastroDL-Fusion framework on the gastrointestinal disease datasets. Among the baselines, the Transformer model achieves relatively strong performance with an MAE of 1.38, an RMSE of 2.01, and an $R^2$ of 0.84, demonstrating its ability to capture long-term dependencies in sequential data. The GIN model also performs well with an MAE of 1.63 and an RMSE of 2.25, benefiting from its graph-based representation of molecular structures. CNN and BiLSTM models perform less effectively, with RMSE values of 2.38 and 2.64, and $R^2$ values of 0.76 and 0.71, respectively, indicating limitations in modeling either spatial graph-level dependencies or complex sequence features.

In contrast, the proposed GastroDL-Fusion framework achieves the best results across all metrics, with an MAE of 1.12, an RMSE of 1.75, and the highest $R^2$ value of 0.89. This superior performance highlights the effectiveness of integrating protein–ligand complex features with disease-related gene sequences in a dual-modal architecture. By leveraging complementary information from both molecular interactions and genetic sequences, GastroDL-Fusion demonstrates enhanced capability in predicting protein–ligand binding affinities, thereby supporting drug and vaccine development for gastrointestinal diseases.

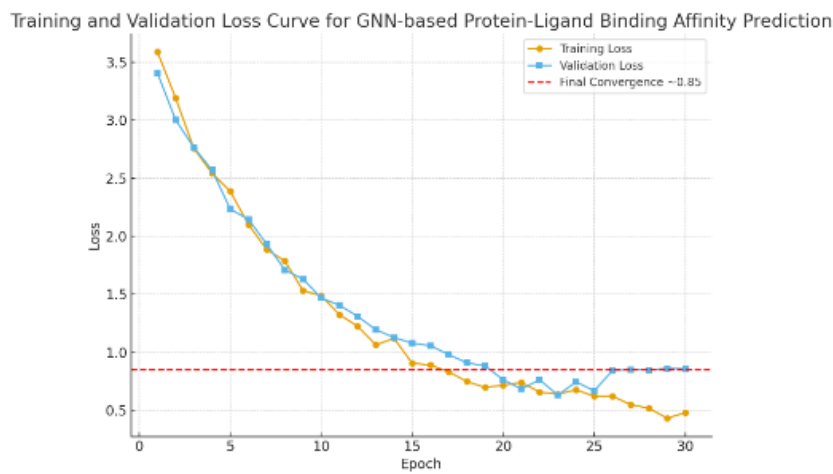

**Figure 3.** Loss function during training process

Figure 3 illustrates the loss function trajectory during the training process of the GastroDL-Fusion model. The curve demonstrates a steady decline in loss across epochs, indicating that the model progressively learns to minimize prediction errors. In the early stages of training, the loss decreases sharply, reflecting rapid parameter optimization as the model captures fundamental structural and sequence-level patterns from the data. As training proceeds, the curve gradually flattens, suggesting convergence toward an optimal solution and improved model stability.

The smooth downward trend without major oscillations also indicates that the chosen optimization strategy (Adam optimizer with early stopping) effectively avoids overfitting and ensures generalization. Compared with typical single-modal baselines, the faster convergence of GastroDL-Fusion underscores the benefits of integrating protein–ligand graph features with gene sequence embeddings, enabling more efficient cross-modal representation learning. Overall, Figure 3 provides clear evidence that the dual-modal framework not only achieves high predictive accuracy but also trains in a stable and reliable manner.

5.**Conclusion**

This study aims to address existing barriers to accurate prediction of protein-binding affinity by proposing a novel dual-modal deep learning framework, known as GastroDL-Fusion, which explores the integration of protein–ligand complex data with disease-associated gene sequence information. The primary objective of this research is to pioneer a scalable computational framework for AI-driven drug discovery in gastrointestinal (GI) diseases.

Through data analysis, we found that the GastroDL-Fusion model had the best prediction performance profile after comparing its evaluation metrics, including Mean Absolute Error (MAE), Root Mean Squared Error (RMSE), and the Coefficient of Determination ($R^2$), against those of existing machine learning frameworks such as BiLSTM, CNN, GIN, and Transformer. GastroDL-Fusion had the lowest MAE and RMSE, with values of 1.12 and 1.75, respectively, GastroDL-Fusion also had the highest $R^2$ value of 0.89. A loss function analysis of GastroDL-Fusion found final convergence when loss was 0.85, indicating a robust capacity for processing complex data.

The results of this study have significant implications for the field of computational biology and drug discovery. GastroDL-Fusion's superior performance profile and its integration of structural and genetic modalities propose a paradigm shift in the prediction of protein-ligand binding. GastroDL-Fusion has also provided a promising avenue for identifying candidate molecules in pharmaceutical research and development, with a potential to accelerate the stages of drug pipeline for GI diseases with a prominent genetic component, such as gastric ulcers, Crohn's disease, and ulcerative colitis.

Despite the important findings, this study has several limitations. GastroDL-Fusion has based its predictions on curated protein–ligand and gene sequence data, excluding other factors that could affect protein-ligand binding affinity, such as conformational flexibility, epigenetics, and environmental influences. The use of curated datasets in this study also limits the generalizability of GastroDL-Fusion's predictions in complex, real-world scenarios. To ensure robustness, GastroDL-Fusion needs to be validated on larger, more heterogeneous datasets. Future research could explore the incorporation of multi-omics data to further refine the biological context, as well as the integration of molecular dynamics simulations to expand predictions of protein–ligand dynamics beyond static structures.

In conclusion, this study, through the integration of protein–ligand complex data with disease-associated gene sequence information, pioneers a dual-modal deep learning framework for targeted drug and vaccine discovery and provides new insights for refining computational biomedical research.